\documentclass[letterpaper, 10 pt, conference]{ieeeconf}  

\IEEEoverridecommandlockouts                              

\usepackage{tikz}
\usetikzlibrary{positioning,fit,arrows.meta,calc}
\usetikzlibrary{matrix, backgrounds}
\usepackage{pgfplots}
\newlength{\figurewidth}
\newlength{\figureheight}
\pgfplotsset{compat=newest}
\usepackage{algorithm}
\usepackage{algpseudocode}
\usepackage{amsmath}
\usepackage{amsfonts}
\usepackage{amssymb}
\usepackage[bookmarks=false,        
            colorlinks=false,       
            pdfborder={0 0 0},      
            breaklinks=true]{hyperref}
\usepackage{mathtools}
\usepackage{xcolor}
\usepackage{booktabs}
\usepackage{color}
\usepackage{graphicx}
\usepackage{rotating}
\usepackage{subcaption}
\usepackage{scalerel}
\usepackage{stackengine}
\newcommand{\ve}[1]{\mathbf{#1}}
\newcommand{\bs}[1]{\boldsymbol{#1}}
\newcommand{\veh}[1]{\ensuremath{\hat{\mathbf{#1}}}}
\newcommand{\bsh}[1]{\ensuremath{\hat{\boldsymbol{#1}}}}
\newcommand{\dve}[1]{\ensuremath{\delta{\mathbf{#1}}}}
\newcommand{\dbs}[1]{\ensuremath{\delta{\boldsymbol{#1}}}}

\newcommand{\Exp}{\operatorname{Exp}}
\definecolor{minty}{RGB}{225,245,238}
\definecolor{lightburgundy}{RGB}{245,228,233}
\definecolor{classygray}{RGB}{240,240,240}
\definecolor{lightcitrus}{RGB}{255, 245, 180}
\definecolor{lightaetnablue}{RGB}{210,225,240}
\definecolor{perkyaetnablue}{RGB}{170,200,250}
\definecolor{evapurple}{RGB}{195, 185, 229}
\definecolor{neogreen}{RGB}{78,255, 74}
\definecolor{evaorange}{RGB}{249, 216, 162}
\definecolor{lavender}{RGB}{197, 202, 233}
\definecolor{coral}{RGB}{255, 213, 194}
\definecolor{paleagua}{RGB}{220, 255, 243}
\definecolor{warmSand}{RGB}{255,248,225}
\usepackage{makecell}
\usepackage{graphicx} 
\usepackage{cite}

\title{\LARGE \bf
Robust Localization for Autonomous Vehicles in Highway Scenes
}

\author{Daqian Cheng$^{1}$, Xuchu Ding$^{2}$, Yujia Wu$^{1}$, Xiang Zhang$^{1}$, and Lei Wang$^{3\dagger}$%
\thanks{*This work was supported by Bot Auto.}%
\thanks{$^{1}$Daqian Cheng, Yujia Wu, and Xiang Zhang are with Bot Auto
{\tt\small {daqian.cheng.162, yujia.wu.365, xiang.zhang.854}@bot.auto}}%
\thanks{$^{2}$Xuchu Ding is with Archer Aviation
{\tt\small xding@archer.com}}%
\thanks{$^{3}$Lei Wang is with Prexa AI
{\tt\small wanglei07@gmail.com}}%
\thanks{$\dagger$Corresponding author}%
}

\begin{document}
\maketitle
\thispagestyle{empty}
\pagestyle{empty}

\begin{abstract}
Localization for autonomous vehicles on highways remains under-explored compared to urban roads, and state-of-the-art methods for urban scenes degrade when directly applied to highways. We identify key challenges including environment changes under information homogeneity, heavy occlusion, degraded GNSS signals, and stringent downstream requirements on accuracy and latency. We propose a robust localization system to address highway challenges, which uses a dual-likelihood LiDAR front end that decouples 3D geometric structures and 2D road-texture cues to handle environment changes; a Control-EKF further leverages steering and acceleration commands to reduce lag and improve closed-loop behavior. An automated offline mapping and ground-truth pipeline keep maps fresh at high cadence for optimal localization performance. To catalyze progress, we release a public dataset covering both urban roads and highways while focusing on representative challenging highway clips, totaling 163 km; benchmarking is standardized using product-oriented accuracy metrics and certified ground truth. Compared to Apollo and Autoware, our system performs similarly on urban roads but shows superior robustness on challenging highway scenarios. The system has been validated by more than one million kilometers of road testing.
\end{abstract}


\section{INTRODUCTION}
Localization is a critical component in autonomous driving systems. Although many methods have been proposed and deployed, most prior work and public benchmarks focus on urban roads; by contrast, highway scenarios, especially long-haul autonomous trucking, remain under-explored. In our experience, when state-of-the-art urban methods are directly migrated to highways, their robustness and availability degrade substantially.


Why is localization different on highways? Based on more than one million kilometers of road testing, we categorize the challenges into external and internal factors. Externally, rapid environmental changes such as road construction and mowing disrupt map–observation consistency, particularly when environment information is highly homogeneous; side-by-side large trucks induce substantial occlusions; and limited mobile network coverage together with sparse Real-Time Kinematic (RTK) service often degrade Global Navigation Satellite System (GNSS) performance. Internally, the truck operates at speeds up to 75 mph (120 km/h), where tracking accuracy becomes highly sensitive to yaw-rate observation errors and delays. Under such conditions, the closed-loop system may exhibit degraded tracking performance or even oscillatory behavior.

Prior localization systems span GNSS \& Inertial Navigation System (INS) integration, map-based LiDAR localization using geometry/intensity/semantics, and visual, inertial multi-sensor fusion. These methods excel on urban roads but rely heavily on stable, repeatable landmarks and, as our experiments indicate, are ill-suited for large-scale environmental changes and severe occlusion. Consequently, highway localization still lacks a robustness-oriented system that explicitly addresses these challenges.

\begin{figure}
    \centering
    \includegraphics[width=1.0\linewidth]{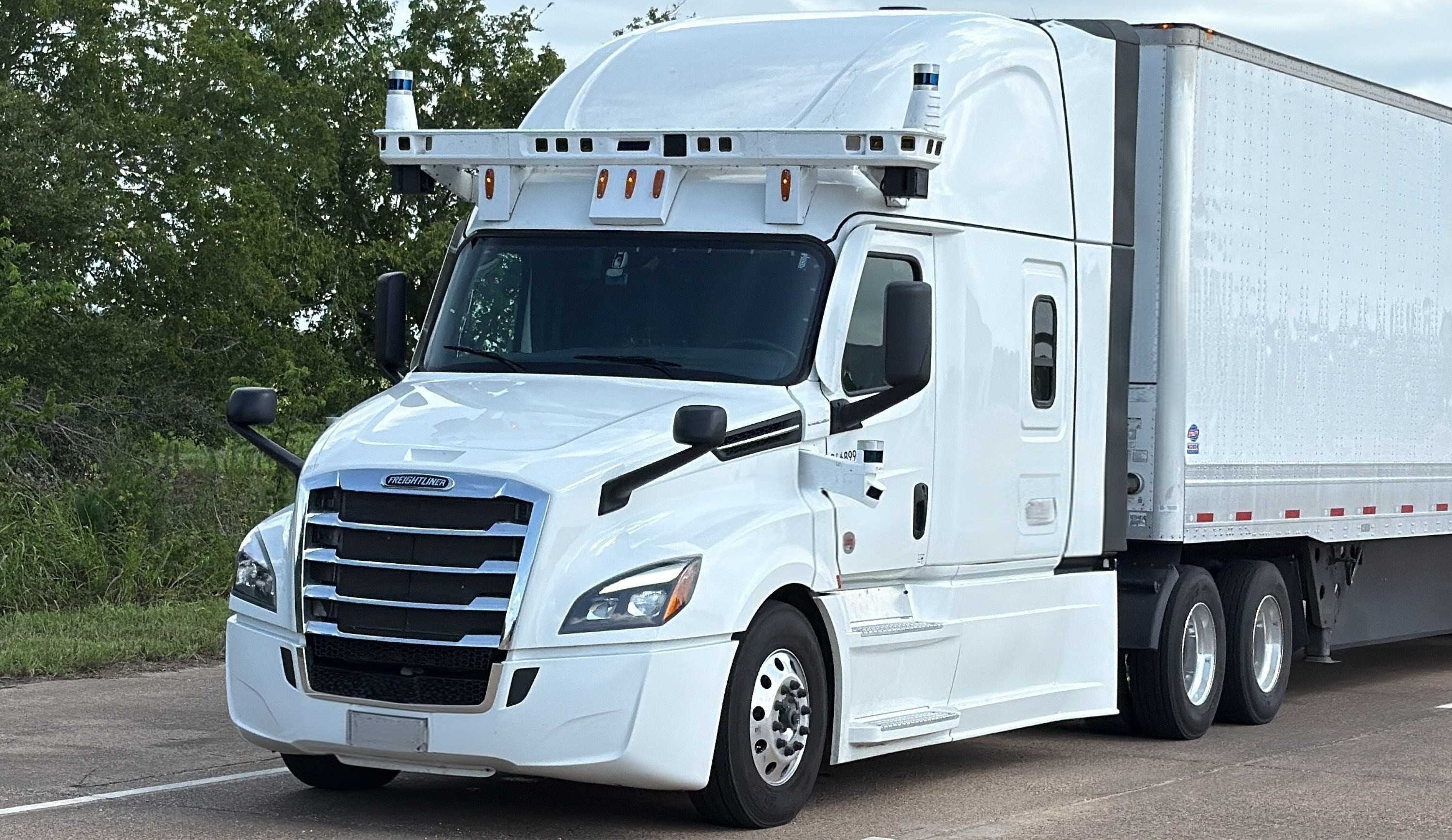}
    \caption{Photo of autonomous truck. Designed for redundancy, the vehicle integrates four long-range 128-line spinning LiDARs (two on top, two on side panels), along with two IMUs and two VCUs installed inside the cabin.}
    \label{fig:truck}
    \vspace{-7mm} 
\end{figure}


To this end, we propose a robust yet simple highway-ready localization system. Environmental information is separated into two modes: 3D geometry (e.g., road structure and terrain shape) and 2D road-surface texture (e.g., lane markings). It is our experience that environmental changes typically affect one mode at a time; therefore, our LiDAR localizer employs two likelihood estimators: one for geometry structure and one for road texture, and fuses them in a Kalman filter, which also provides complementary robustness under heavy occlusion. An automated mapping pipeline keeps maps fresh, updating them daily to ensure map-observation inconsistencies are addressed; rare cases where geometry and texture change simultaneously are considered out of scope without materially harming availability. To improve tracking performance and mitigate oscillatory behavior, we introduce a Control EKF that integrates steering and acceleration commands into a data-driven truck model, reducing localization lag and aligning estimates with real dynamics. Despite these enhancements, the system remains simplistic with modest hardware requirements. Compared to Apollo \cite{wan2018msf} and Autoware \cite{autoware_software}, which both perform well in typical conditions, our system shows superior robustness under challenging highway scenarios.

To promote community progress, we release a public dataset focusing on highway challenges: construction, mowing, and heavy occlusion, alongside typical highway and urban segments for baseline comparison\footnote{Dataset will be made publicly available on Bot Auto website (bot.auto).}. Representative clips are visualized in Fig. \ref{fig:case_study}. The benchmark includes 48 clips covering 163 km. Evaluation is standardized with certified ground-truth trajectories and product-oriented accuracy metrics, detailed in Sec. \ref{subsec:dataset}.

We summarize our contributions as follows:
\begin{itemize}
\item \textbf{A robust localization system for highway challenges.} The system withstands environmental change and heavy occlusion, and achieves low lag for high-speed control by integrating vehicle control signals. This localization system has been proven through 1M km of road testing.
\item \textbf{Highway localization dataset \& evaluation protocol.} We identify key highway-specific challenges and release representative clips, together with standardized metrics and certified ground truth.
\end{itemize}


\section{RELATED WORK}
\label{sec:related}

Modern localization for autonomous driving primarily adopts multi-sensor fusion frameworks that integrate GNSS/INS, LiDAR matching, and vision feature alignment. For LiDAR matching, existing methods can be broadly divided into geometry-based and learning-based approaches. Geometry-based LiDAR localization methods \cite{wan2018msf, wolcott2017gmm, pan2021tusen, li2022tightly} typically register online scans to prebuilt maps using variants of Iterative Closest Point (ICP) or Normal Distributions Transform (NDT) \cite{biber2003ndt}, as well as voxel- or grid-based likelihood estimations.
Learning-based approaches, such as \cite{li2021l3net, michel2018lidarintensity}, directly regress poses from raw or projected LiDAR measurements, demonstrating robustness against geometry degradation and perceptual aliasing. More recently, EgoVM \cite{he2023egovm} estimates poses via cross-modality matching between BEV perception features and lightweight vector maps, employing a robust histogram solver.
For vision-based localization, feature alignment methods leverage image representations. Approaches such as PoseNet \cite{kendall2015posenet}, hourglass network regressors \cite{meyer2018hourglass}, LSTM-based correlation models \cite{walch2017lstm}, and geometry-aware loss formulations \cite{kendall2017geomloss} have enabled image-based 6 degree of freedom (DoF) localization. Among these, geometry-based LiDAR methods remain the most widely deployed in real-world systems, proving to be more robust. Furthermore, some localization frameworks \cite{jo2011interacting_control_ekf, ma2019ack_control_ekf, min2019kinematic_control_ekf} incorporate control commands to enhance fusion accuracy.

Generic LiDAR odometry methods—including KISS-ICP \cite{vizzo2023kissicp}, CT-ICP \cite{dellenbach2022cticp}, LIO-SAM \cite{shan2020liosam}, LOAM \cite{zhang2014loam}, and FAST-LIO2 \cite{xu2022fastlio2}—have achieved impressive odometry accuracy on public datasets. However, adapting these methods for production-level localization requires additional effort, as they must address odometry drift that compromises global consistency.

Compared to urban settings, research on highway localization is relatively limited. A comprehensive survey \cite{laconte2021highway_survey} highlights existing approaches primarily designed for driver-assistance systems, which fall short of meeting the stringent accuracy and robustness requirements of L4 autonomy. Other works rely on lane markings for localization \cite{harr2018highway_loc, choi2019higway_loc}, but such methods are vulnerable to lane repavement and environmental changes.

Public localization benchmarks have significantly advanced the field, but remain biased toward urban environments. KITTI \cite{geiger2012kitti} continues to serve as a canonical odometry and localization benchmark, yet it contains limited highway scenarios. Pit30m \cite{martinez2020pit30m} and Oxford RobotCar \cite{maddern2017robotcar} provide larger-scale datasets but are still urban-centric. Apollo SouthBay \cite{lu2019l3} includes one freeway segment, but is located in a metropolitan area that is comparatively feature-rich and does not represent U.S. interstate corridors. In addition, this segment lacks the challenging highway scenarios explored in our work. These gaps motivate the development of our highway-centric dataset, which emphasizes challenging scenarios for robust localization.

\section{Localization Method}
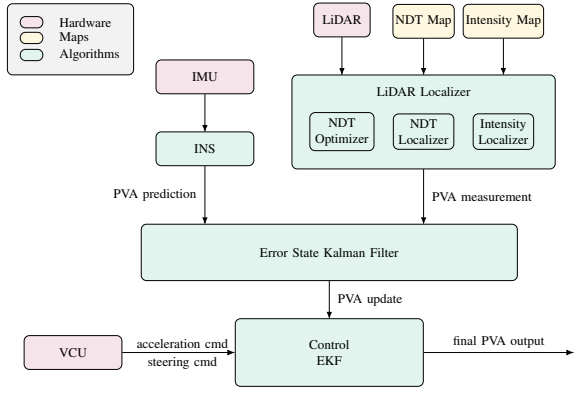
\begin{figure}[tp]
    \centering
    \scalebox{0.5}{

\begin{tikzpicture}[
  >=Latex,
  node distance=10mm and 15mm,
  solidbox/.style={draw, rounded corners, minimum width=26mm, minimum height=9mm, align=center},
  dashedbox/.style={draw, dashed, rounded corners, minimum width=26mm, minimum height=9mm, align=center},
  bigbox/.style={draw, rounded corners, minimum width=100mm, minimum height=15mm, align=center},
  group/.style={draw, rounded corners, inner sep=6pt},
  arrow/.style={->, line width=0.6pt},
  darr/.style={->, dashed, line width=0.6pt}
]

\node[solidbox, fill = lightburgundy] (imu) {IMU};
\node[solidbox,  fill=minty, below=of imu] (insp) {INS};
\draw[->] (imu.south) -- (insp.north);


\node[bigbox, fill = minty, minimum width=100mm, below=20mm of insp.east, xshift=20mm] (eskf) {Error State Kalman Filter};


\node[group,  fill = minty, right=10mm of imu, yshift=-12mm, minimum width=70mm, minimum height=25mm, label={[yshift=-7mm]90:LiDAR Localizer}] (lidarloc) {};

\node[solidbox, inner sep=4pt, minimum width=15mm] (ndloc) at ($(lidarloc.south)+(0,10mm)$) {NDT \\ Localizer};
\node[solidbox, inner sep=4pt, minimum width=15mm,left=5mm of ndloc] (ndtopt) {NDT \\ Optimizer};
\node[solidbox, inner sep=4pt, minimum width=15mm, right=5mm of ndloc] (intloc) {Intensity \\ Localizer};

\node[solidbox, fill = lightburgundy, above=of ndtopt, yshift=10mm, inner sep=2pt, minimum width=15mm] (lidar) {LiDAR};
\node[solidbox,fill=warmSand, above=of ndloc, yshift=10mm, inner sep=2pt, minimum width=15mm] (ndtmap) {NDT Map};
\node[solidbox,fill=warmSand,above=of intloc, yshift=10mm, inner sep=2pt, minimum width=15mm] (intmap) {Intensity Map};

\draw[->] (lidar.south) -- ($(lidar.south)!(lidarloc.north)!(lidar.south |- lidarloc.north)$);
\draw[->] (ndtmap.south) -- ($(ndtmap.south)!(lidarloc.north)!(ndtmap.south |- lidarloc.north)$);
\draw[->] (intmap.south) -- ($(intmap.south)!(lidarloc.north)!(intmap.south |- lidarloc.north)$);

\draw[->] (insp.south) -- ($(insp.south)!(eskf.north)!(insp.south |- eskf.north)$)node[midway,  left, xshift=-1mm] {PVA prediction};

\draw[->] (lidarloc.south) -- ($(lidarloc.south)!(eskf.north)!(lidarloc.south |- eskf.north)$)node[midway,  right, xshift=1mm] {PVA measurement};

\node[solidbox,fill = minty, minimum width=50mm, minimum height = 18mm, below=10mm of eskf] (controlekf) {Control \\ EKF};
\node[solidbox, fill = lightburgundy, left = 30mm of controlekf](vcu){VCU};

\draw[->] (eskf.south)  -- ($(eskf.south)!(controlekf.north)!(eskf.south |- controlekf.north)$)node[midway,  right, xshift=1mm] {PVA update};
\draw[arrow] (vcu.east) -- (controlekf.west) node[midway, above, xshift=1mm] {acceleration cmd} node[midway, below, xshift=1mm] {steering cmd};
\draw[arrow] (controlekf.east) -- ++(40mm,0)
    node[midway, above] {final PVA output};

\tikzset{
  legend box/.style={draw, rounded corners, fill=gray!10, inner sep=6pt},
  legend swatch/.style={draw, rounded corners, minimum width=6mm, minimum height=3mm}
}

\matrix (legend) [matrix of nodes,
                  nodes={anchor=west, inner sep=1pt},
                  row sep=2pt, column sep=4mm,
                  left=8mm of imu, yshift=10mm] {
  \node[legend swatch, fill=lightburgundy] {};   & Hardware \\
  \node[legend swatch, fill=warmSand] {};& Maps \\
  \node[legend swatch, fill=minty] {};         & Algorithms \\
};

\begin{scope}[on background layer]
  \node[legend box, fit=(legend)] {};
\end{scope}

\end{tikzpicture}}
\caption[All blocks with lightcitrus denote hardware.]%
 {Localization system overview. LiDAR localizer outputs pose measurement and covariance, ESKF fuses INS prediction and LiDAR measurement, and Control EKF consumes control commands plus ESKF output to produce final PVA estimation.
}
    \label{fig:OverallSystem}
    \vspace{-5mm}
\end{figure}

Our localization framework estimates the autonomous vehicle's position, velocity, and attitude (PVA) jointly under the Universal Transverse Mercator (UTM) coordinate system. A standard Error State Kalman Filter (ESKF) fuses the 100 Hz PVA prediction from INS based on IMU readings and the 10 Hz pose (PA) estimation from the LiDAR Localizer using the 3D and intensity information of the LiDAR point cloud. Finally, to make the localization result smooth and low-lag enough for downstream, a Control EKF module takes the acceleration and steering command from Vehicle Control Unit (VCU) as prediction and takes the PVA estimation from ESKF as measurement to produce the final localization result. See Fig. \ref{fig:OverallSystem} for our overall system design. We do not utilize the GNSS system other than for bootstrapping, because we find that it often fails sporadically due to limited RTK coverage and mobile network services.

We will cover the LiDAR Localization algorithms, the ESKF framework and the Control EKF in \ref{lidar_localizor}, \ref{eskf_framework} and \ref{control_ekf}, respectively. 

\subsection{LiDAR Localizer} \label{lidar_localizor}
LiDAR Localizer matches the pattern of onboard LiDAR point cloud with pre-built maps to provide pose estimations. We build the following two maps offline using the GNSS and LiDAR data collected:
\begin{itemize}
    \item An NDT map which voxelizes the 3D space into $0.8\, m$ cubes; each voxel stores the 3D mean and covariance of the LiDAR points' positions within the voxel.
    \item An intensity map which partitions the space into a horizontal 2D grid of $0.125\,m$ cells; each cell stores the mean and variance of the LiDAR intensities within.
\end{itemize}
Please find the details of our mapping process in \ref{mapping}.

Our LiDAR localization algorithm is a Maximum Likelihood Estimator (MLE). The idea is to sample poses around a given prior provided by ESKF prediction, then compute the likelihood of each sampled pose based on the geometry information from the NDT map and the texture information from the intensity map, and output the pose with the maximum likelihood.  To speed-up runtime, sample pose evaluations are performed on high-performing Graphics Processing Units (GPU) in parallel.

Algo. \ref{lidar_localization_algorithm} summarizes our LiDAR localization algorithm. For notation, we use $\ve{p}$,  $\ve{R}$ to denote the UTM position and rotation matrix $\mathrm{SO}(3)$. We use $Z$ to denote the point cloud obtained onboard aggregated from all LiDARs at the localization time. Each point has an intensity value and a 3D position $\ve{q}$ in the body frame.

\begin{algorithm} [H] 
\caption{LiDAR Localizer}\label{lidar_localization_algorithm} 
\textbf{Inputs:} Prior pose $ (\veh{p}, \veh{R})$; NDT map $M_{NDT}$; Intensity map $M_{Int}$; onboard LiDAR point cloud $Z$\\
\vspace{-3mm}
\begin{algorithmic}
\State $\ve{p}^*, \ve{R}^* = \text{InitialOptimization}(\veh{p}, \veh{R}, M_{NDT}, Z) $
\State $\{(\ve{p}^1, \ve{R}^1), ..., (\ve{p}^N, \ve{R}^N)\} = \text{PoseSample}(\ve{p}^*, \ve{R}^*) $ 
\State $Z_{NDT} = \text{SampleForNDT}(Z)$
\State $Z_{Int} = \text{SampleForIntensity}(Z)$
\For{ i = 1, 2, ..., N}
  \State $L^i \leftarrow 0$
  \State $L^i \leftarrow L^i + \text{NDTLogLikelihood}(\ve{p}^i, \ve{R}^i, M_{NDT}, Z_{NDT})$
  \State $L^i \leftarrow L^i + \text{IntensityLogLikelihood}(\ve{p}^i, \ve{R}^i, M_{Int}, Z_{Int})$
  \State $\text{Prob}^i \propto e^{-L^i}$
\EndFor
\State $\ve{p}, \ve{R}, \ve{W} = \text{MeanAndCovariance}(\{\ve{p}^i\}, \{\ve{R}^i\}, \{\text{Prob}^i\})$
\State \Return $\ve{p}, \ve{R}, \ve{W}$
\end{algorithmic}
\end{algorithm}
\vspace{-3mm}

\subsubsection{Initial Optimization}
Although the pose-wise likelihood calculations are executed on GPU, sampling poses on all six DoF is computationally intractable.
Since the z, pitch, and roll can be readily and accurately estimated by LiDAR-map alignment because of the presence of the ground plane, our strategy is to decouple the 6 DoF: first use a lightweight optimizer to estimate z, pitch, and roll, then use MLE for the x, y, and yaw dimensions.

In Algo. \ref{lidar_localization_algorithm}, InitialOptimization is the algorithm to estimate z, pitch and roll. It is an NDT method that iteratively improves pose estimation. We sample $M$ LiDAR points  $Z_{NDT}=\{\ve{q}^1,..., \ve{q}^M\}$ from the onboard point cloud $Z$. In each iteration $k$, we transform the sampled points $\ve{q}^j$ into the UTM frame denoted by $\ve{u}^j = \ve{R}_{k-1} \ve{q}^j + \ve{p}_{k-1}$ using the pose $(\ve{p}_{k-1}, \ve{R}_{k-1})$ from the previous iteration. We look up the voxel that $\ve{u}^j$ falls in and get the corresponding 3D mean $\bs{\mu}^j$ and covariance $\bs{\Sigma}^j$ stored in the voxel in the NDT map $M_{NDT}$. Levenberg-Marquardt algorithm is used to iteratively optimize the pose estimation $\ve{p}_k, \ve{R}_k$ z, pitch, roll components to maximize the log likelihood: 
\begin{equation}
-\sum_j (\ve{R} \ve{q}^j + \ve{p} - \bs{\mu}^j)^T (\bs{\Sigma}^j)^{-1} (\ve{R} \ve{q}^j + \ve{p} - \bs{\mu}^j) 
\end{equation}
\vspace{-2mm}


\subsubsection{Pose and Point Cloud Sampling}
After the initial pose optimization step, we sample $N$ candidate poses $(\ve{p}^1, \ve{R}^1), ..., (\ve{p}^N, \ve{R}^N)$ around the optimized pose $\ve{p}^*, \ve{R}^*$ to be used for MLE. In our implementation  we choose $N$ to be 8000. We fix the $z$, pitch, and roll dimensions and sample the $x,y$, yaw dimensions uniformly with sampling range $0.5\,m, 1.0\,m, 0.5^\circ$ respectively, which achieves a good balance between convergence radius and sample efficiency.

To compute the NDT likelihood and intensity likelihood, we need to downsample the raw point cloud $Z$ to reduce computation workload. Two different sampling strategies are used because the NDT evaluation captures the environment's geometry pattern which has more macroscale consideration compared to intensity evaluation which captures more microscale texture information.
Therefore, for the NDT likelihood calculation, we first use a $(1m)^3$ sized 3D voxel-grid sampler to ensure a sparse and uniform sampling and then randomly down-select 5000 points to ensure computation efficiency. For the intensity likelihood calculation, we employ a $(0.1m)^2$ sized 2D grid sampler followed by a 30000 point random down-selection.


\subsubsection{NDT and Intensity Likelihood}
To calculate the NDT likelihood, we use the 5000 point samples $Z_{NDT}$. Let $\ve{p}, \ve{R}$ denote the pose being evaluated. We transform each point sample $\ve{q}^j \in Z_{NDT}$ to the UTM frame as $\ve{u}^j = \ve{R} \ve{q}^j + \ve{p}$. We look up the voxel that $\ve{u}^j$ falls in and get the corresponding 3D mean $\bs{\mu}^j$ and covariance $\bs{\Sigma}^j$ stored in the voxel in the NDT map $M_{NDT}$. We calculate the log likelihood as the pose $\ve{p}, \ve{R}$ as
\begin{equation}
-\sum_j (\ve{u}^j - \bs{\mu}^j)^T (\bs{\Sigma}^j)^{-1} (\ve{u}^j - \bs{\mu}^j) 
\end{equation}
\vspace{-2mm}

The intensity likelihood computation is similar. We just replace the 3D voxel matching by the 2D grid matching and use the intensity mean and variance in each matched grid for likelihood computation. 

At last, we compute a distribution $\{\text{Prob}^i\}$ where $\text{Prob}^i \propto e^{-L^i}$ for the sampled poses,  and return the mean $\ve{p}, \ve{R}$ and covariance $\ve{W}$ (under Lie algebra) of those poses as the final output of the LiDAR Localizer.

\subsection{ESKF Framework} \label{eskf_framework}
The localization state is denoted by $\ve{x}=(\ve{p}, \ve{v}, \bs{\theta})$ where $\ve{p}, \ve{v}$ are the UTM position and linear velocity; $\bs{\theta}$ is the rotation vector of the attitude. We also use the rotation matrix representation $\ve{R}$ interchangeably with $\bs{\theta}$ to represent the attitude.  

The error state is denoted by $\dve{x}  = (\dve{p}, \dve{v}, \dbs{\theta})$ and it is defined as $\dve{x} := \ve{x} - \veh{x}$, 
where $\ve{x}$ and $\veh{x}$ are the true and nominal states respectively. 

In ESKF, we need to predict the nominal state and the error state; and we need measurement to update the error state and correct the nominal state by the error state.  
\subsubsection{INS} \label{InsPredictor}
INS uses the following system dynamic equations for nominal state prediction at time $k$:
\begin{equation} \label{eq:ins_predictor}
    \begin{split}
        \veh{p}_k =& \ve{p}_{k-1} + \ve{v}_{k-1}\Delta t \\ 
        \veh{v}_k = & \ve{v}_{k-1} + (\ve{R}_{k-1}\tilde{\ve{a}}  + \ve{g})\Delta t\\
        \veh{R}_k = & \ve{R}_{k-1}\Exp(\tilde{\bs{\omega}} \Delta t)\\ 
    \end{split}
\end{equation}
Here $\Delta t$ is the time difference between the time step $k$ and $k-1$; and $\tilde{\ve{a}}, \tilde{\bs{\omega}}$ are the acceleration and angular velocity readings of the IMU in the body frame. For simplicity, we assume that the IMU biases have already been corrected. 

The dynamic of the error state can easily be derived from the above system dynamics:
\begin{equation} \label{error_state_dynamic}
    \begin{split}
        \hat{\dve{p}}_k = & \dve{p}_{k-1} + \dve{v}_{k-1} \Delta t \\ 
        \hat{\dve{v}}_k = &  \dve{v}_{k-1}  -[(\ve{R}_{k-1}\tilde{\ve{a}} ) \times  \dbs{ \theta}_{k-1}]\Delta t\\
        \hat{\dbs{\theta}}_k = & \Exp(-\tilde{\bs{\omega}} \Delta t) \dbs{\theta}_{k-1} \\ 
    \end{split}
\end{equation}
We apply standard error state covariance matrix prediction based on \ref{error_state_dynamic}; details are omitted for brevity.
 
\subsubsection{ESKF Update} \label{eskf_update}
LiDAR Localizer takes the ESKF's prediction as input and returns the its pose estimation and covariance matrix as the measurement update in ESKF. Therefore, the measurement model is $\ve{z}= h(\ve{x}) = (\ve{p}, \boldsymbol{\theta})$. Note that the Jacobian of $h$ in ESKF is defined as
\begin{equation}
\ve{H} = \left. \frac{\partial h(\veh{x} + \dve{x} )}{\partial \dve{x}} \right |_{\dve{x} = 0}
\end{equation}
where $\veh{x}$ is the nominal state (predicted state) and $\dve{x}$ is the error state.

Using Lie algebra (see \cite{Barfoot2017}), we have 
\begin{equation} 
\left. \frac{\partial \text{Log} (\text{Exp}(\bsh{\theta})\text{Exp}(\dbs{\theta})}{\partial \dbs{\theta}}\right |_{\dbs{\theta}=0} = \ve{J}_r^{-1}(\bsh{\theta})
\end{equation}
Therefore, in our case,  
\begin{equation} \label{H_matrix}
\ve{H}(\veh{x}) = 
\begin{bmatrix}
 \ve{I}_3 & \ve{0}_3 & \ve{0}_3  \\
 \ve{0}_3 & \ve{0}_3 & \ve{J}_r^{-1}(\boldsymbol{\hat{\theta}}) \\
\end{bmatrix}
\end{equation}
where $\ve{I}_3$ and $\ve{0}_3$ are the $3\times 3$ identity and zero matrices. We apply the standard ESKF update using \eqref{H_matrix}.

\subsection{Control EKF} \label{control_ekf}

\label{ControlPredictor}
Control EKF is used to smooth the ESKF output to meet the smoothness requirement posted by the downstream control module. Unlike a conventional INS-based EKF that is driven directly by sensor outputs, the Control EKF uses steering and acceleration commands as inputs. This accounts for the intrinsic lags introduced by combustion and chassis dynamics, enabling the prediction step to better capture vehicle behavior and improve estimation accuracy.

A physics based truck model is inherently complicated, even under ideal assumptions \cite{chen2000lateral, chen1997vehicle}. We use {\em SINDy} \cite{brunton2016discovering} to capture a set of simple governing equation for the normal driving data. The resulting dynamics can be expressed in the discrete-time state-space form,
\begin{equation}
\begin{split}\label{eq:control_discrete}
    x_{k+1} =& x_{k} - v_{k} \sin(\phi_{k})\Delta t, \\  
    y_{k+1} =& y_{k} +  v_{k}\cos(\phi_{k})\Delta t, \\ 
    v_{k+1} =&v_{k} +  a_{k}\Delta t, \\ 
    \phi_{k+1} = & \phi_{k} +  r_{k}\Delta t, \\
    r_{k+1} = & r_{k} +  [(w_1/v_{k} +w_2v_{k}) r_{k} + w_3 \sigma^{\text{cmd}}_{k}]\Delta t, \\ 
    a_{k+1} = & a_{k} + w_4 (-a_{k}+ a^{\text{cmd}}_{k})\Delta t ,
\end{split}
\end{equation}
where $(x_k, y_k)\in\mathbb{R}^2$ denote the UTM position, $v_k \in \mathbb{R}$ is the vehicle speed, $a_k \in \mathbb{R}$ the longitudinal acceleration, $\phi_k \in \mathbb{R}$ the yaw angle in the UTM frame, and $r_k \in \mathbb{R}$ the yaw rate. The system inputs are the steering angle $\sigma^{\text{cmd}}_k \in \mathbb{R}$ and the commanded acceleration $a^{\text{cmd}}_k \in \mathbb{R}$. 
We assume the process disturbances are applied to control inputs. Based on this state-space model \eqref{eq:control_discrete}, the design of the Extended Kalman Filter follows directly from the standard textbook formulation.

\section{Ground Truth Generation and Mapping} \label{mapping}

In this section, we describe the offline processes to generate localization ground-truth (GT) trajectory for each of our trips without relying on high-cost GT-purpose systems. Our GT framework is based on Simultaneous Localization and Mapping (SLAM). It serves two purposes. One is to generate localization ground-truth for general trips which enables our fully automated pipeline to validate localization in these trips.  The other is to generate mapping ground truth for candidates trips for high-definition (HD) mapping. Reliable mapping infrastructure is a critical prerequisite for robust highway-grade autonomous localization.  In this section we also summarize the design strategies and implementation details of our mapping pipeline.

A key feature of ground-truth and mapping framework is that it is fully automated and low cost, which enables us to build HD maps in high cadence and be highly scalable as we scale up our fleet.  Another is that we do not fully depend on GNSS systems for our ground-truth framework (i.e., GNSS data are used only when available).  This is useful in the case of GNSS outages, which can occur due to bad weather, structural occlusions, signal interference, weak coverage, and a multitude of other factors that can arise in practice as we accumulate many miles of driving data. 

We use one consistent framework to generate both types ground-truth mentioned above, with the only difference being how the SLAM system is set up.  We use a well-known factor graph framework to organize constraints.  A factor graph provides a flexible representation for fusing heterogeneous sensor modalities and constraints into a unified optimization problem, and have been widely used in state-of-the-art SLAM backends~\cite{dellaert2017factor}.  In this framework, robot poses are modeled as nodes in the graph, while constraints are represented as factors connecting these nodes. The Maximum A Posteriori (MAP) estimate of the trajectory is obtained by solving a nonlinear least-squares problem over this graph structure. Ceres Solver \cite{agarwal2022ceres} is used to model and solve the MAP optimization problem.

\begin{figure}
    \centering
    \includegraphics[width=1.0\linewidth]{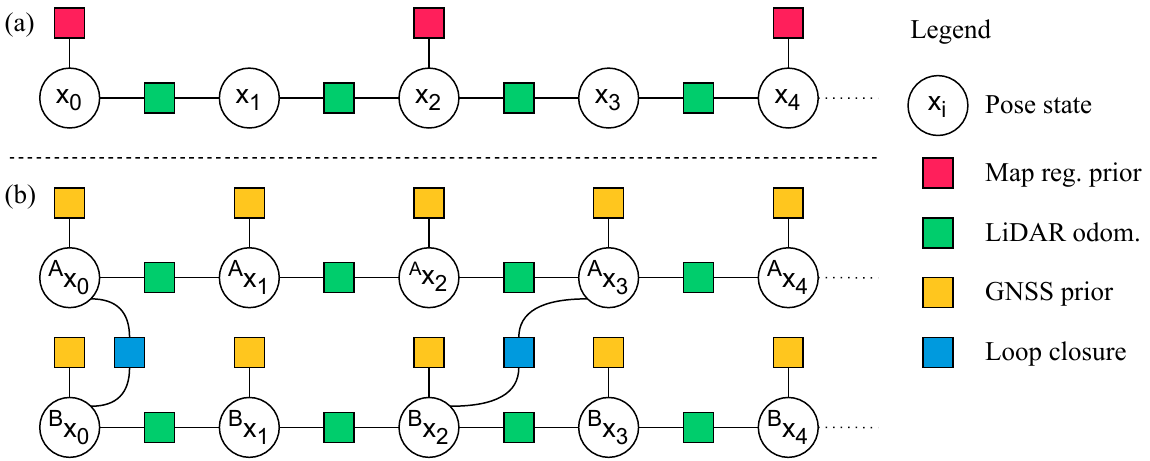}
    \caption{Factor-graphs of the proposed ground-truth framework. (a) shows localization GT formulation, where map prior factors are obtained using point cloud registration with the trip's corresponding mapping trips. (b) shows mapping GT formulation, where loop closures are generated between two or more overlapping mapping sessions.}
    \label{fig:factor_graph}
    \vspace{-6mm} 
\end{figure}

Fig. \ref{fig:factor_graph} illustrates the factor graph design for localization and mapping GT generation.  For localization GT, we treat the map as reference, as the goal of localization is to produce a trajectory estimate under which the registered onboard point cloud aligns well with the mapping point cloud. In our implementation, we form prior factors using G-ICP \cite{akita2022smallgicp} point cloud registration between onboard point cloud frames and a dense offline LiDAR point cloud map derived from the mapping process. We employ KISS-ICP LiDAR odometry \cite{vizzo2023kissicp} to generate odometry factors between each two consecutive frames to ensure local smoothness. The localization GT has superior accuracy to onboard LiDAR localization by leveraging future information and a full-density point cloud map. 

For mapping, the goal is to generate trajectories that produce complete, consistent, and sharp point cloud maps. Therefore, to produce one map, we use more than one trips to alleviate occlusion-induced map incompleteness of any single pass. All overlapping mapping sessions must be aligned to produce one consistent point cloud map; therefore, we form loop closure factors among co-visible mapping sessions via G-ICP point cloud registration. We employ the same LiDAR odometry factor to ensure local point cloud consistency. Global accuracy is ensured by GNSS prior factors (if they are available).


\section{Experiments}

\subsection{Dataset with Diverse Highway Challenges}
\label{subsec:dataset}

We identify three major external challenges for localization on highway: road construction, terrain shift, and occlusion. Highway environments can be highly homogeneous; on some stretches, lane markings are the only texture cue and the terrain elevation is the only 3D geometry structure. Therefore, environment change like roadwork and terrain shift can induce
rapid and extensive map-observation inconsistencies. Specifically, road repavement may shift lane markings, misleading lane marking-based localization methods; road-side grass mowing can drastically reduce terrain elevation by 1 meter. Heavy occlusion by side-by-side trucks can block LiDAR's field of view, causing degraded localization accuracy and sometimes degeneration. Some clips are presented in detail in Sec. \ref{subsec:localization_benchmark}.

\begin{figure*}[htbp]
    \centering
    \includegraphics[width=\textwidth]{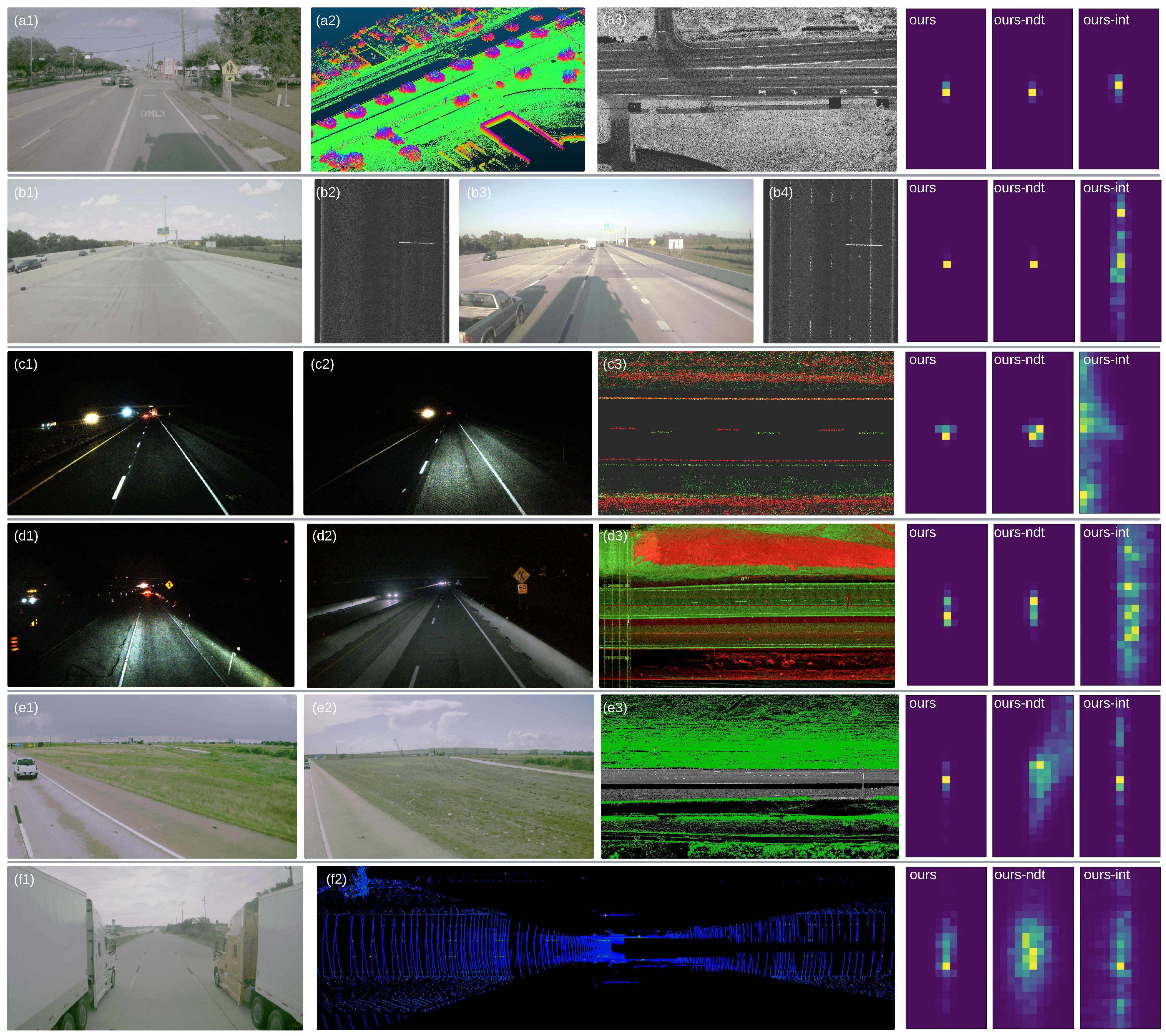}
    \caption{Case study on urban road and challenging scenarios. All cases show a sample frame's LiDAR localizer distribution marginalized to the lateral (±0.5m) and longitudinal (±1.0m) dimensions, including NDT-only and intensity-only variants. (a) shows an urban road segment; LiDAR point clouds colored by height (a2) and by intensity (a3, in bird's-eye view) show abundant geometric and road texture information respectively. (b) A repavement from Botts' Dots (b1, b2) to painted markings (b3, b4) makes intensity localizer highly uncertain, and fusion relies on NDT localizer. (c) A repavement with shifted painted markings in both lateral and longitudinal direction (c3, from red to green) results in a biased and uncertain intensity localization, but NDT remains accurate and confident due to no geometric change. (d) A repavement with road deviation; the road surface was widened with large structure change (d3, from red to green), resulting in insufficient intensity information match; but NDT remains reliable thanks to terrain and unchanged road structures (e.g., the overhead bridge). (e) Roadside mowing shifts terrain elevation and affects NDT because the grass terrain can occupy most of LiDAR scan, as shown in the LiDAR semantic segmentation visualization (e3), while intensity localization helps keep fusion result accurate. (f) When ego is "sandwiched" by two other trucks, LiDAR's view becomes heavily occluded (f2, in bird's-eye view), causing uncertain NDT localization; but intensity localization helps maintain accurate and certain lateral localization, which is safety-critical in this scenario.}
    \label{fig:case_study}
\end{figure*}

We release a highway-centric dataset collected on autonomous trucks in Fig. \ref{fig:truck}. The dataset consists of 48 mapping-testing clip pairs, including 9 urban and 39 highway segments. Instead of providing lengthy but unremarkable recordings, 23 of the highway clips are deliberately selected to capture challenging scenarios, comprising 9 construction zones, 7 terrain shifts, and 7 occlusions. Fig. \ref{fig:case_study} visualizes some challenging clips. The remaining 16 clips are categorized as typical highway, representing general highway conditions. The dataset totals 163 km and collected on Interstate highway corridors in Texas, U.S., which are some of the most freight-concentrated corridors. The dataset has diverse weather conditions, seasons, and time of day.

The evaluation is standardized with product-oriented metrics. In contrast to academic convention of evaluating statistical accuracy, we prioritize robustness and deem performance as pass-or-fail. Specifically, standard highway lane width and truck width dictates a lateral accuracy of 20 cm; empirically, downstream object detection module requires a yaw accuracy of 0.3° and a longitudinal accuracy of 50 cm in order to correctly associate detection with HD-map, e.g., a larger yaw error may result in incorrect vehicle-lane associations.

\subsection{Localization Benchmark}
\label{subsec:localization_benchmark}

We compare the proposed localization method, including NDT-only and intensity-only variants, to two public and open-source baselines, Apollo \cite{wan2018msf} and Autoware \cite{autoware_software}. Although more recent methods exist, these two frameworks have been widely deployed in real-world and exhibit adequate accuracy and strong robustness. Both baselines share similar LiDAR localization and EKF framework design as ours. For LiDAR localization, Autoware uses an NDT variant; Apollo mainly utilizes the LiDAR intensity information via a 2-D intensity map, but also weakly incorporates geometric information using a 2-D elevation map. In contrast, our method takes advantage of geometry information more heavily with a 3D voxelized NDT map. For a fair comparison, we utilize Apollo's built-in option to only use GNSS for bootstrapping, which is consistent with Autoware and our methods. Our method achieves 46 ms mean runtime on AMD EPYC 7643 CPU and RTX 4090 GPU.

Both baselines are tuned for optimal performance adaptation to our platform. In particular, we find Autoware less production-ready for highway operation for the following reasons: (1) Autoware uses dense LiDAR point clouds as map, which demands impractically large onboard storage space for our 400 km+ long routes; (2) as the vehicle travels fast, dynamically loading large dense map chunks becomes a bottleneck; (3) native FP32 map point cloud precision does not accommodate long routes and subsequently degrades localization accuracy. For the experiments, we bypassed these limitations by slow-playing the dataset and pre-shifting the map origin to eliminate the floating-point precision issues.

All five methods pass the urban road dataset as shown in Table \ref{tab:exp_loc_surface}, showing that both Apollo and Autoware generalize well to our truck platform. The performance of ours-ndt and ours-intensity demonstrates that the surface roads contain abundant geometric and texture information, and accurate localization can be achieved by utilizing either information. In Table \ref{tab:exp_loc_typical_highway}, we show the results for the typical highway dataset.  It can be seen that methods that utilize 3D geometric information perform well, namely ours, ours-ndt, and autoware; whereas methods that rely on texture, i.e., Apollo and ours-intensity, exhibit large longitudinal errors in some cases. The distribution visualizations (see Fig. \ref{fig:case_study}) suggest intensity-based LiDAR localizations have large longitudinal uncertainty, showing that the texture information in highway environments is less reliable than geometric information in these cases, especially to provide longitudinal localization constraints. 


\begin{table}[h]
  \centering
  \renewcommand{\arraystretch}{1.2}
  \vspace{-2mm}
  \caption{Accuracy evaluation on urban road dataset}
  \begin{tabular}{l|r r r r r r}
    \Xhline{1.2pt}
    Method & \thead{lat/m\\RMSE} & \thead{lat/m\\Max}  & \thead{lon/m\\RMSE} & \thead{lon/m\\Max} & \thead{yaw/°\\RMSE}  &  \thead{yaw/°\\Max} \\
    \hline
    Ours     & 0.019 & \textbf{0.071} & 0.032 & 0.137 & \textbf{0.050} & 0.282 \\
    Ours-ndt & 0.019 & 0.083 & 0.029 & 0.137 & 0.051 & \textbf{0.280} \\
    Ours-int & 0.019 & 0.073 & 0.028 & 0.169 & \textbf{0.050} & 0.285 \\
    Apollo   & 0.034 & 0.154 & 0.071 & 0.254 & 0.079 & 0.289 \\
    Autoware & \textbf{0.016} & 0.076  & \textbf{0.020} & \textbf{0.087} & 0.064 & 0.284 \\
    \Xhline{1.2pt}
  \end{tabular}
  \label{tab:exp_loc_surface}
  \vspace{-3mm}
\end{table}

\begin{table}[h]
\centering
\caption{Accuracy evaluation on typical highway dataset}
\setlength{\tabcolsep}{3pt} 
\renewcommand{\arraystretch}{1.2}

\begin{tabular}{l | r r | r r r r r r}
\Xhline{1.2pt}
Method & \thead{Clip\\Fail} & \thead{Frame\\Fail} &
\thead{lat/m\\RMSE} & \thead{lat/m\\Max} &
\thead{lon/m\\RMSE} & \thead{lon/m\\Max} &
\thead{yaw/°\\RMSE} & \thead{yaw/°\\Max} \\
\hline
Ours     & \textbf{0} & \textbf{0\%} & \textbf{0.022} & 0.179 & 0.031 & \textbf{0.135} &  0.038 & \textbf{0.259} \\
Ours-ndt & \textbf{0} & \textbf{0\%} & 0.025 & \textbf{0.169} & \textbf{0.027} & 0.144 &  \textbf{0.037} & 0.262 \\
Ours-int & 4/15 & 0.14\% & \textbf{0.022} & 0.184 & 0.067 & 0.653 &  0.046 & 0.276 \\
Apollo   & 3/15 & 1.70\% & 0.067 & 0.297  & 0.108 & 0.948 & 0.118 & 1.548 \\
Autoware & \textbf{0} & \textbf{0\%} & \textbf{0.022} & 0.179 & 0.031 & 0.246  & 0.050 & 0.273 \\
\Xhline{1.2pt}
\end{tabular}
\label{tab:exp_loc_typical_highway}
\vspace{-2mm}
\end{table}

For road construction, we find repavements to have the greatest impact on localization by mostly affecting intensity LiDAR localizer. Experiment shows geometric-reliant methods remain robust against repavements. Different types of repavements have different effects on intensity-reliant methods. For Botts' Dots, intensity localizers show high uncertainty and subject to large error due to the sparsity and LiDAR blooming. Repavements with shifted lane markings can mislead localization. Some road work would guide the vehicle to deviate from the original road, such as road widening projects, which would cause insufficient information and render intensity localization ineffective, often leading to loss of localization (see Fig \ref{fig:case_study}-b,c,d).

Rapid terrain elevation shifts are usually caused by roadside mowing. Although geometric LiDAR localizers are often robust to this scenario, we find them occasionally uncertain and error-prone, and could lead to loss of localization if not handled. Since road texture is unaffected, intensity localizer can help stabilize LiDAR localization, as shown in Fig. \ref{fig:case_study}-e. In practice, we also mitigate the problem by maintaining fresh maps and by segmenting out affected terrain map.

Side-by-side large trucks cause LiDAR occlusion and reduces information for localization. When occluded on one side, geometric LiDAR localization remains unaffected. However, when occluded on both sides, i.e., ego is ``sandwiched" by two other trucks, we find the geometric and texture information to be complementary:  the geometric localizer becomes uncertain, but the intensity localizer provides lateral constraint, as shown in Fig. \ref{fig:case_study}-f. Lateral accuracy and certainty is safety critical when ego is driving between two other trucks. Another type of occlusion involves both mapping and onboard: the mapping dataset was occluded on one side, while onboard localization is occluded on the other side; this scenario exhibits similar characteristics to the ``sandwich'' occlusion cases.  The results for our method and Apollo / Autoware in the above three types of challenging environments are summarized in Table \ref{tab:exp_loc_challenging}.

\begin{table}[h]
  \centering
  \caption{Pass/Fail comparisons on the challenging dataset}
  \begin{tabular}{l|c c c}
    \Xhline{1.2pt}
    Method & \thead{Repavement (9)} & \thead{Terrain Shift (7)} & \thead{Occlusion (7)}  \\
    \hline
    Ours     & \textbf{0} & \textbf{0} & \textbf{0} \\
    Apollo   & 4 & 2 & 5 \\
    Autoware & 1 & 2 & 1 \\
    
    \Xhline{1.2pt}
  \end{tabular}
  \label{tab:exp_loc_challenging}
  \vspace{-5mm}
\end{table}


\subsection{Evaluation of the Control EKF Module}

\begin{figure}[t]
    \centering
    \scalebox{0.5}{\input{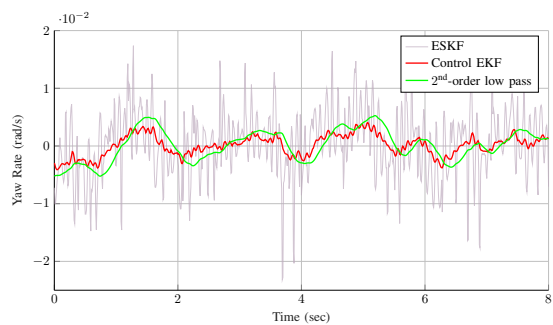}}
\caption[Control EKF result comparison]%
 {Yaw rate comparison on highway. The raw ESKF output is too noisy for downstream control module, and using a low pass filter introduces lag. The proposed control EKF achieves smoother yaw rate estimation without lag.}
    \label{fig:control_experiment}
    \vspace{-5mm}
\end{figure}

From \eqref{eq:control_discrete}, it is evident that the lateral feedback loop is highly sensitive to yaw-rate estimation noise. A second-order low-pass filter is typically used to reduce noise, with a cost of introducing lag in estimation. Field tests show that when yaw rate delay exceeds 100 ms at 120 km/h, the closed-loop system has little stability margin and can oscillate with small disturbances. We evaluate 3 yaw-rate signals: the direct output from ESKF, the Control EKF, and a second-order low-pass filter. In the ESKF, the INS runs at 100 Hz and unavoidably propagates IMU noise, producing a noisy yaw-rate output. The Control EKF, in contrast, embeds vehicle dynamics into its prediction step, which suppresses noise and reduces latency. Fig. \ref{fig:control_experiment}  illustrates the difference: the ESKF output is dominated by high-frequency noise, while the Control EKF yields a smoother estimate and reduces effective delay by $\sim$100 ms than the low-pass filter. This improvement directly benefits closed-loop tracking.

\section{CONCLUSIONS}
In this work, we highlight the unique challenges of localization on highways compared to urban roads. Through extensive comparisons with baseline methods in challenging highway scenarios, we demonstrate that leveraging geometric structural information is essential for reliable highway localization, while incorporating road texture cues offers complementary robustness under particularly difficult conditions. Moreover, integrating control commands into the localization framework enhances trajectory smoothness without introducing additional system lag. Finally, our public dataset addresses a critical gap in existing localization benchmarks, and we believe it will enable more comprehensive evaluation and foster future research on robust localization.













\bibliographystyle{IEEEtran}
\bibliography{ref}


\end{document}